\documentclass[conference]{IEEEtran}
\IEEEoverridecommandlockouts
\usepackage{cite}
\usepackage{amsmath,amssymb,amsfonts}
\usepackage{algorithm}
\usepackage{algorithmic}
\usepackage{tabularx}
\usepackage{graphicx}
\usepackage{textcomp}
\usepackage{xcolor}
\usepackage{booktabs}

\newcommand{\etal}{\mbox{\emph{et al.\ }}}

\def\BibTeX{{\rm B\kern-.05em{\sc i\kern-.025em b}\kern-.08em
    T\kern-.1667em\lower.7ex\hbox{E}\kern-.125emX}}
\begin{document}

\title{Graph Neural Networks for IceCube Signal Classification \\
\thanks{\IEEEauthorrefmark{7}See http://icecube.wisc.edu/collaboration/authors/current for full author list and acknowledgments.}
}
\author{
\IEEEauthorblockN{
Nicholas Choma\IEEEauthorrefmark{1}, 
Federico Monti\IEEEauthorrefmark{2}, 
Lisa Gerhardt\IEEEauthorrefmark{3}, 
Tomasz Palczewski\IEEEauthorrefmark{4}, 
Zahra Ronaghi\IEEEauthorrefmark{3}, 
Prabhat\IEEEauthorrefmark{3}, 
Wahid Bhimji\IEEEauthorrefmark{3},\\
Michael M. Bronstein\IEEEauthorrefmark{2}\IEEEauthorrefmark{5}, 
Spencer R. Klein\IEEEauthorrefmark{4}\IEEEauthorrefmark{6},
Joan Bruna\IEEEauthorrefmark{1}
for the IceCube collaboration\IEEEauthorrefmark{7}
}
\and
\IEEEauthorblockA{\textit{\IEEEauthorrefmark{1}Courant Institute of Mathematical Sciences} \\
\textit{New York University}\\
New York NY 10012 USA}
\and
\IEEEauthorblockA{\textit{\IEEEauthorrefmark{2}Institute of Computational Science} \\
\textit{Universit\`a della Svizzera italiana}\\
Lugano 6900, Switzerland}
\and
\IEEEauthorblockA{\textit{\IEEEauthorrefmark{4}Department of Physics}\\
\textit{University of California, Berkeley}\\
Berkeley CA 94720 USA}
\and
\IEEEauthorblockA{\textit{\IEEEauthorrefmark{3}NERSC} \\
\textit{Lawrence Berkeley National Laboratory} \\
Berkeley CA 94720 USA}
\and
\IEEEauthorblockA{\textit{\IEEEauthorrefmark{5}Department of Computing} \\
\textit{Imperial College}\\
London SW7 2AZ, UK}
\and
\IEEEauthorblockA{\textit{\IEEEauthorrefmark{6}Nuclear Science Division} \\
\textit{Lawrence Berkeley National Laboratory}\\
Berkeley CA 94720 USA}
\and
\IEEEauthorblockA{
nc2201@cims.nyu.edu, 
federico.monti@usi.ch, 
lgerhardt@lbl.gov, 
tpalczewski@berkeley.edu,
zronaghi@lbl.gov,
prabhat@lbl.gov,\\
wbhimji@lbl.gov,
m.bronstein@imperial.ac.uk,
srklein@lbl.gov,
bruna@cims.nyu.edu}
}

\maketitle

\begin{abstract}
Tasks involving the analysis of geometric (graph- and manifold-structured) data have recently gained prominence in the machine learning community, giving birth to a rapidly developing field of geometric deep learning. 
In this work, we leverage graph neural networks to improve signal detection in the IceCube neutrino observatory.
The IceCube detector array is modeled as a graph, where vertices are sensors and edges are a learned function of the sensors' spatial coordinates. 
As only a subset of IceCube's sensors is active during a given observation, we note the adaptive nature of our GNN, wherein computation is restricted to the input signal support.
We demonstrate the effectiveness of our GNN architecture on a task classifying IceCube events, where it outperforms both a traditional physics-based method as well as classical 3D convolution neural networks.
\end{abstract}


\section{Introduction}

Graph- and manifold-structured data are ubiquitous in a broad range of research fields studying interactions and relations between objects. Social networks, molecules, point clouds, and 3D shapes are just a few of many different examples of such data. The remarkable success of deep learning techniques in computer vision has motivated in recent years the development of deep learning architectures capable of coping with data defined on non-Euclidean domains, referred to by the general term {\em Geometric Deep Learning} \cite{bronstein2017geometric}. 

A plethora of geometric deep learning models have been proposed, including 
learnable information diffusion processes \cite{gori2005new,scarselli2009graph,li2016gated,gilmer2017neural}, 
generalizations of convolutional neural networks (CNN) \cite{lecun1998gradient} to graphs using 
spectral filters 
\cite{bruna2013spectral,henaff2015deep,defferrard2016convolutional,kipf,levie2017cayleynets,monti2017geometric,monti2018motifnet} 
and local neighborhood operations   
\cite{duvenaud2015convolutional,monti2016geometric,atwood2016diffusion,hamilton2017inductive,velivckovic2017graph,wang2018dynamic}. 
Geometric deep learning has been successfully employed in a broad spectrum of applications, ranging from 
computer graphics, vision  \cite{masci2015geodesic,boscaini2016learning,wang2018dynamic}, and  medicine \cite{parisot2017spectral,zitnik2018modeling} to chemistry \cite{duvenaud2015convolutional,gilmer2017neural} and  high energy physics \cite{henrionneural}. 

In this paper, we study the application of graph neural networks  (GNN) to the challenging problem of neutrino detection in the IceCube observatory. 
One of the key motivations for the use of GNNs is the irregular geometry of the detectors, as well as the  sparse nature of the signal recorded by the IceCube detectors. 
A significant challenge setting apart the neutrino detection problem from typical pattern classification problems is a very large asymmetry between positive and negative events; requiring good accuracy in the regime of extremely low False Positive Rates on the order of $10^{-6}$.

\subsection{IceCube Experiment}

\begin{figure}[tbp]
\centerline{\includegraphics[scale=0.25]{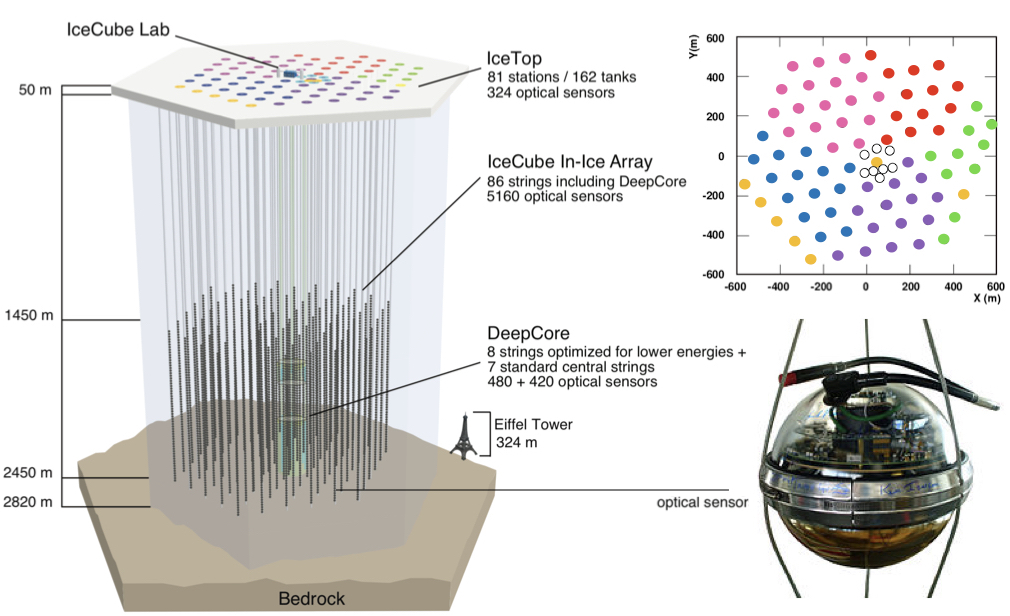}}
\caption{The IceCube Neutrino Observatory with the in-ice array, its sub-array DeepCore, and the cosmic-ray air shower array IceTop. The string color scheme represents different deployment seasons. The top-right insert presents the top view of the IceCube detector. The DeepCore sub-array is represented by open circles. 
}
\label{fig:detector}
\end{figure}

IceCube is a 1 km$^3$ neutrino observatory located at the South Pole \cite{Abbasi:2008aa}.  
Its primary purpose is to look for high-energy (above 100 gigaelectronvolts (GeV)) neutrinos that are produced by the same cosmic particle accelerators that produce ultra-high energy cosmic-rays \cite{Halzen:2010yj}.   
Its 5,160 sensors (digital optical modules, or DOMs) detect the Cherenkov light that is produced by the relativistic charged particles resulting from high-energy neutrinos interacting in the Antarctic ice.   
The Cherenkov light is emitted at an angle and may scatter before being observed by sensors that are typically 10 to 60 meters away from the track.  
Sixty sensors are deployed on each of 86 strings placed in holes drilled in the ice. 
Most of the strings are on a 125 m triangular grid, but 8 strings, forming the 'Deep Core' infill array, have much tighter spacing. 
On most strings, the DOMs are deployed every 17 m, from 1450 m to 2450 m below the surface; in Deep Core, most of the DOMs are deployed with a 7 m spacing between 2100 and 2450 m. 
IceCube includes an array of 81 surface stations called IceTop, designed to study cosmic ray interactions in the atmosphere. 
The schematic view of the IceCube detector is shown in Fig.~\ref{fig:detector}. 
The sensors record the photon arrival times using waveform digitizers.  Across the array, the relative arrival times are known to better than 3 ns \cite{Abbasi:2008aa}.

IceCube observes two classes of events. Contained events occur when neutrinos interact within the detector.  Through-going events are mostly long-lived muons which can travel many kilometers in the ice.  They can be produced in neutrino interactions, or, in downward-going events, in cosmic-ray air showers that occur when high-energy cosmic-rays interact in the upper atmosphere.  

IceCube has observed a spectrum of astrophysical neutrinos from energies of about 35 teraelectronvolts (TeV) to above 2 petaelectronvolts (PeV) \cite{Aartsen:2014gkd,Aartsen:2016xlq}.  At lower energies, there is an overwhelming background of neutrinos produced when cosmic-rays interact in our atmosphere.  At higher energies, several factors emerge.  First, the neutrino-nucleon cross-section rises and the Earth becomes opaque to neutrinos, so we cannot observe neutrinos coming from much below the horizon \cite{Aartsen:2017kpd}.  Second, the flux drops sharply at higher energies. 

\begin{figure}[tbp]
\centerline{\includegraphics[scale=0.77]{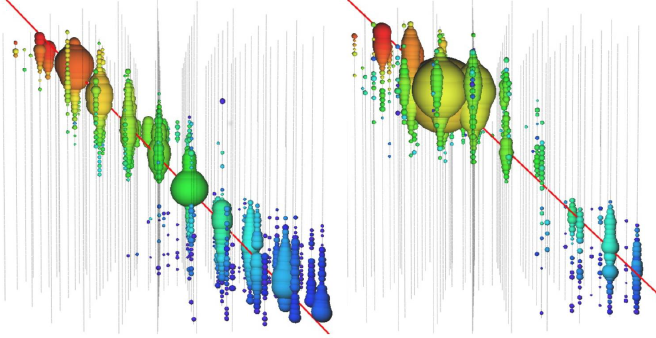}}
\caption{The characteristic pattern of light deposition for muon bundles (left) and a high-energy single muon with visible stochastic light emission along the track (right). The red line shows the actual (Monte Carlo track), while each colored bubble represents a DOM that saw light in the event.  The colors indicate the relative light arrival time, from red (earliest) to blue (latest), while the size of the bubbles indicates the number of observed photons.}
\label{fig:stochasticity}
\end{figure}

A study of downward-going muons from astrophysical neutrino interactions could significantly expand IceCube's reach in this area.  Such a search is challenging because of the very high flux of muons from cosmic-ray air showers.  In this work, we discuss methods to separate this signal (muons from neutrinos) from the background (muons from cosmic-ray showers).  For the present purposes, the main difference between the signal and the background is the stochasticity of the energy deposition, which translates into clumpiness in the light emission from the track.  Muons from neutrinos are single muons, which lose energy stochastically, leading to a very uneven light emission along the muon track.   In contrast, muons from cosmic-ray interactions come in bundles containing from one (rarely) to hundreds of muons.  With the large number of muons, the light emission averages out, and becomes quite smooth (see Fig.~\ref{fig:stochasticity}).  There is still variation in light detection due to the depth-dependent optical properties of the ice. 

We generated two Monte Carlo data sets, one for signal and one for background, and used them for all three methods discussed in this paper. In this work, we consider only astrophysical muon neutrino events as a signal. In our signal simulation set, neutrino energies range from 100~GeV to 10$^{8}$~GeV. The astrophysical neutrino data set was assumed to follow a power law, with $dN_\nu/dE_\nu \propto E_\nu^{-2.0}$, while the cosmic-ray air shower background was generated with the CORSIKA simulation package \cite{Heck:1998vt} using H4a composition assumption \cite{Gaisser:2012}. The energy range of cosmic ray primaries in our background simulation set is 600 GeV to 10$^{5}$~GeV. No other backgrounds are taken into consideration.  For both signal and background, more energetic events are likely to pass the event selection.  So, to efficiently generate our samples, we generated the signal and background samples following a harder spectral index ({\it i. e.} biased toward more energetic events).  Each event has an associated weight, so that weighted histograms reproduce the output from the original spectrum.  This allows considerably more efficient use of computer resources, but it becomes harder to evaluate the statistical significance of any result.

The particles produced by these simulations were then run through an IceCube-specific detector simulation, which included the generation of Cherenkov light, light propagation through the ice, and the detector response. For uniformity, the DeepCore array and IceTop stations were excluded during the preselection process. The resulting simulated data were run through the standard IceCube calibration and reconstruction packages.  
To be usable by IceCube, the final event sample must have a reasonably high signal-to-noise ratio (SNR).  Here, to compare different methods, we decided a-priori to evaluate the methods on the basis of how many events they could find, subject to maintaining a 1:1 SNR.  Since the background is many orders of magnitude larger than the signal, this requires a very high level of rejection.

\section{Baseline Architectures}

\subsection{Physics Baseline Method}

IceCube has developed two conventional ({\it i. e. not machine learning}) criteria to measure stochasticity and reject muon bundles.  The first divides the muon track into 120 m long segments, and determines the light output from the track in each segment \cite{Abbasi:2012wht}.  This light output is then fit to a line, and a pseudo-$\chi^2$ as the sum of squared residuals is calculated. Events with large $\chi^2$ are very stochastic.  The second approach also reconstructs the light deposition along the track, by examining the light output in 50 meter long segments, apportioning light from each DOM to the nearest segment.  Then, the energy loss in the largest segment is divided by the mean energy loss, giving the peak/median ratio.  The results from these two methods are correlated, but the correlation is low enough that there is value in using both. IceCube uses a combination of hand-tuned selections on these two variables to select single muon events. The baseline method utilizes stochasticity selections and selections in energy proxy versus zenith phase space to limit the cosmic muon background contribution \cite{TP:neutrino2018}.

\subsection{3D Convolution Neural Network}
The geometry of the IceCube detector is shown in Figure~\ref{fig:detector}. 
Convolution neural networks typically perform convolution on orthogonal coordinate systems, so the input data is transformed and mapped onto an orthogonal grid \cite{Mirco:2017}. We mapped 86 strings of the detector onto a 10x20 grid and hexagonally shaped input was transformed to orthogonal by padding with zeros. Each string of the detector holds 60 DOMs, as a result input tensor with three spatial dimensions (10x20x60) is fed into a 3D CNN with 18 layers. ResNet-18 \cite{He:2016} is a 18 layer convolution neural network, where the residual learning framework and shortcut connections ease the training of deep networks. These residual blocks consist of convolutions and a summed shortcut connection to the output; convolutions were done using Keras framework with Tensorflow backend. Deepcore strings were added with strings 86, 81, 82 on one row and strings 85, 79, 84, 83, 80 on another row.  We extracted information from simulated background and signal data by aggregating events from hdf5 files events (collection of information from the sensors with a fixed time window).

\section{Graph Neural Networks}
\subsection{Overview}

{

First attempts of learning on graphs constructed features resulting from the steady state of a learnable diffusion process \cite{gori2005new,scarselli2009graph}. This approach has recently been improved with modern deep learning schemes using gated recurrent units \cite{li2016gated} and neural message passing \cite{gilmer2017neural}. 


Bruna \etal \cite{bruna2013spectral,henaff2015deep}  proposed formulating convolution-like operations in the spectral domain, defined by the eigenvectors of the graph Laplacian operator. 
The key idea is that Laplacian eigenvectors form an orthogonal basis allowing performing Fourier decomposition of graph-based signals; filtering is thus performed by multiplying the Fourier coefficients of the input signal by learnable spectral representation of a filter. 
%
%
%
%
Among the drawbacks of this architecture is $\mathcal{O}(n^2)$ computational complexity due to the cost of computing the forward and inverse graph Fourier transform (multiplication by the dense matrix of eigenvectors), 
$\mathcal{O}(n)$ parameters per layer, and no guarantee of spatial localization of the filters.

A more efficient class of spectral graph CNNs was introduced in \cite{defferrard2016convolutional,kipf2016semi} and follow up works, who proposed spectral filters that can be expressed in terms of simple operations (such as additions, scalar- and matrix multiplications) w.r.t. the Laplacian. In particular, Defferrard \etal \cite{defferrard2016convolutional} considered Chebyshev polynomial filters of degree $p$, which require only $p$ multiplications by the Laplacian matrix (each such multiplication costs $\mathcal{O}(n)$ assuming that the graph is sparsely connected). Due to the local nature of the Laplacian operator, such filters are guaranteed to have a $p$-hop support. 
Levie \etal \cite{levie2017cayleynets} extended this framework using rational filter functions that also include inversions of the Laplacian, which are carried out approximately using an iterative Jacobi method. 
Monti \etal used multivariate polynomials w.r.t. multiple Laplacians defined by graph motifs \cite{monti2018motifnet} as well as Laplacians defined on products of graphs \cite{monti2017geometric}.

A different class of methods are spatial formulations of graph CNNs, which operate on local neighborhoods on the graph  
\cite{duvenaud2015convolutional,monti2016geometric,atwood2016diffusion,hamilton2017inductive,velivckovic2017graph,wang2018dynamic}. 
Monti \etal \cite{monti2016geometric} proposed the Mixture Model networks (MoNet), generalizing the notion of image `patches' to graphs. The centerpiece of this construction is a system of local pseudo-coordinates $\mathbf{u}_{ij} \in \mathbb{R}^d$ assigned to a neighbor $j$ of each vertex $i$. The spatial analogue of a convolution is then defined as a set of weighting kernels applied in these coordinates. In particular, when  Gaussian kernels are used, the graph convolution can be interpreted as a Gaussian mixture, 
\begin{equation} \label{eq:monet}
\tilde{\mathbf{x}}_i = \mathrm{ReLU} \left( \sum_{m=1}^M y_m \sum_{j \in \mathcal{N}_i} e^{-(\mathbf{u}_{ij}-\boldsymbol{\mu}_m)^\top \boldsymbol{\Sigma}^{-1}_m (\mathbf{u}_{ij}-\boldsymbol{\mu}_m)} \mathbf{x}_j \right),
\end{equation}
where $\mathbf{x}_i$, $\tilde{\mathbf{x}}_i$ denotes the $q'$- and $q$-dimensional input and output feature column vectors at vertex $i$, respectively, $\mathcal{N}_i$ is the neighborhood of $i$,  and $y_1, \hdots, y_M$,  
$\boldsymbol{\mu}_1, \hdots, \boldsymbol{\mu}_M$ and $\boldsymbol{\Sigma}_1, \hdots, , \boldsymbol{\Sigma}_M$ are the learnable Gaussian mixture parameters (mixture weights, mean vectors, and covariance matrices, respectively).  
Each Gaussian applied to the input features can be considered as one dimension of the local $M$-dimensional patch (in the Euclidean setting, one pixel).

Veli\v{c}kovi\`c \etal \cite{velivckovic2017graph} reinterpreted this scheme as {\em graph attention} (GAT), learning the relevance of neighbor vertices for the filter result, 
\begin{equation} \label{eq:monet_layer}
	\tilde{\mathbf{x}}_i = \mathrm{ReLU}\left(\sum_{j \in \mathcal{N}_i} \alpha_{ij} \mathbf{x}_j\right),
\end{equation}
where 
\begin{equation} \label{eq:gat_layer}
\alpha_{ij} = \frac{
\exp(\mathrm{LeakyReLU}(\mathbf{b}^\top[\mathbf{A}\mathbf{x}_i,\, \mathbf{A}\mathbf{x}_j])
}
{
\sum_{k \in \mathcal{N}_i} \exp(\mathrm{LeakyReLU}(\mathbf{b}^\top[\mathbf{A}\mathbf{x}_i,\, \mathbf{A}\mathbf{x}_j] )
}
\end{equation}
are attention scores representing the importance of vertex $j$ w.r.t. $i$, 
and the $p \times q'$ matrix $\mathbf{A}$ and $2p$-dimensional vector $\mathbf{b}$ are the learnable parameters.  
The graph attention mechanism was extended in \cite{kondor2018n} using high-order tensor covariant models, in \cite{bruna2017community} using the non-backtracking operator defined on the line graph, and in \cite{monti2018dual} using alternating graph convolutions with dual graph convolutions.
}

\subsection{Graph neural networks on IceCube}

The IceCube experiment can be considered as a geometric domain given by the $x,y,z$ coordinates of each DOM, resulting in a 3D irregular hexagonal grid. 
For a given event, active DOMs output three real-valued numbers (representing the energy and the time of the interaction), which can be thought of as a signal defined on a fixed graph. 
Since many DOMs are inactive, such a signal is very sparse. Alternatively, one can consider each event as a signal on a  different graph, containing only the active DOMs. 
Graph neural networks, capable of dealing both with irregular geometry and graphs of different size, thus appear perfect candidates for the IceCube experiment. 
Further, a prior assumption typically made on data when using convolutional neural networks is that of stationarity, due to the common assumption of shift-invariance in e.g. natural images \cite{lecun1998gradient}. 

{Because the IceCube detector array is hexagonal and irregular, stationarity is a too strong assumption in our case as DOMs may present different distances and orientations one w.r.t. the other. This would involve very fine-grained grids and consecutively large filters or many layers to extract meaningful hierarchical features from the provided input signal, potentially requiring vast amounts of parameters and thus leading to overfitting.} 
Graph neural networks do not require this type of stationarity, as distances between pairs of DOMs can be represented with graph edges which are calculated using a learned kernel function. {The flexibility that graph operators introduce in the model allows in this sense to realize rich filters in space while varying just a significantly small amount of parameters (e.g. in the MoNet framework presented above, we just need to change the mean and standard deviations of the considered Gaussian kernels for analyzing different regions of space). This naturally allows to implement significantly smaller models, improving generalization over unseen conditions and ultimately performance at test time.}

\section{Network Architecture}
\label{sec:arch}


\paragraph{Data}
  The input to our network is an {\em event}, consisting of $n$ DOMs represented as graph vertices. 
To each input vertex, a 6-dimensional feature vector is associated, containing the $x,y,z$ position of its corresponding DOM, the sum of charge in the first pulse within the sensor, the sum of charge in all pulses within the sensor, and the time at which the first pulse crosses the activation threshold. 
The task is to classify events into positives (neutrino) or negatives (background). 
A typical event activates only a sparse set of DOMs (see example in Fig.~\ref{fig:stochasticity}); the GNN architecture naturally allows to handle such situations by ignoring the inactive vertices, considering graphs of different size. 
%

\paragraph{Graph construction}
The geometry of the detector is represented by a fixed, weighted, directed graph. The weights are defined only by the spatial positions ($x,y,z$) of the DOMs. 
The $n\times n$ adjacency matrix $\mathbf{A}$ of the graph is constructed similarly to \cite{monti2016geometric} by applying a Gaussian kernel to the pairwise distances between the DOMs, followed by the softmax operation, 
\begin{align}
	d_{ij} &= e^{-\frac{1}{2}\|\mathbf{x}_{i}-\mathbf{x}_{j}\|^2 / \sigma^2} \\
    a_{ij} &= \frac{e^{d_{ij}}}{\sum_k e^{d_{ik}}},
\end{align}
where $n$ is the number of DOMs (vertices of the graphs), $\mathbf{x}_i$ denotes the spatial coordinates of the $i$th DOM, and $\sigma$ is a learnable scalar parameter controlling the locality of the kernel and how fast the matrix $\mathbf{A}$ spreads information across distant vertices. 


\paragraph{Graph convolution}
We apply a sequence of $T$ graph convolutional layers to the input data, as follows. Each layer $t$ inputs $d^{(t)}$-dimensional feature vector at each vertex arranged into an $n\times d^{(t)}$ matrix $\mathbf{X}^{(t)}$ and outputs $d^{(t+1)}$-dimensional features represented as an $n\times d^{(t+1)}$ matrix $\mathbf{X}^{(t+1)}$. 

The graph convolution is performed by applying the adjacency matrix to the input 

\begin{align}
\mathrm{GConv}(\mathbf{X}^{(t)}) = [\mathbf{A} \mathbf{X}^{(t)}, \,\,  \mathbf{X}^{(t)}](\mathbf{a}^{(t)})^\top + b^{(t)}\boldsymbol{1}, 
\end{align}
where the learnable parameters are a $2d^{(t+1)}$-dimensional vector of weights $\mathbf{a}^{(t)}$ and the scalar bias $b^{(t)}$, different for each layer. 
The output of the layer is produced by concatenating the result of the graph convolution with a non-linear activation thereof, 
\begin{align}
\mathbf{X}^{(t+1)} = [\mathrm{ReLU}(\mathrm{GConv}(\mathbf{X}^{(t)}), \,\, \mathrm{GConv}(\mathbf{X}^{(t)})],
\end{align}
which was experimentally found beneficial to achieve better performance.


\paragraph{Pooling}
The output features of the $T$th graph convolutional layer are pooled using a sum operation over the $n$ vertices,
\begin{align}
	x_k^{(pool)} &= \sum_{i=1}^n x^{(T)}_{ik}, \quad k=1,\hdots, d^{(T)}
\end{align}
producing a $d^{(T)}$-dimensional vector $\mathbf{x}^{(pool)}$, to which 
logistic regression is applied to output the predicted class of the event,
\begin{align}
    \hat{y} &= \mathrm{sigmoid}((\mathbf{x}^{(pool)})^\top \mathbf{a}^{(pool)} + b^{(pool)}),
\end{align}
where $\mathbf{a}^{(pool)}$ is a $d^{(T)}$-dimensional vector of weights and $b^{(pool)}$ is the scalar bias parameter of the output layer.


\section{Experiment and Results}

\subsection{Training and Evaluation}
Model training, validation, and testing were completed using a signal and a background dataset containing 25250 and 109491 events, respectively.
Each dataset was then subdivided into 50\% training, 25\% validation, and 25\% test sets. 
We performed early stopping once performance was maximized during training on the validation set, up to 100 epochs.
Of our trained models, we selected for final evaluation the model which performed best on the test set.

Final model evaluation was performed using additional signal and background datasets containing 8487 and 366433 events, respectively.

Each event is assigned a weight during its generation corresponding to the event's frequency of occurrence.
As such, all training and evaluation was performed on weighted samples, keeping in line with the goal of selecting as many signal events per year as possible while maintaining a 1:1 SNR.

Table \ref{table:dataset} displays the number of events for both the signal (SG) and background (BG) classes within each dataset, with and without event weighting.
The number of weighted events for a given class within a given dataset is the sum of all weights of events in the dataset which belong to that class.
Thus while the class imbalance is large for the unweighted samples, it is far larger once event weighting is applied.

\begin{table}[htbp]
 \caption{Unweighted and weighted number of signal and background events within each dataset}
\label{table:dataset}
\begin{tabular}{l*{4}{r}}
  & \multicolumn{2}{c}{\# Unweighted} & \multicolumn{2}{c}{\# Weighted} \\
  \cmidrule(lr){2-3} \cmidrule(lr){4-5}
  Dataset & Signal &  Background & \multicolumn{1}{r}{Signal} & Background \\ 
  \midrule
  Training & 12624 & 54745 & 7.8 & 37275 \\
  Validation & 6313 & 27373 & 3.9 & 18648 \\
  Test & 6313 & 27373 & 3.9 & 18632 \\
  Final Evaluation & 8487 & 366433 & 5.2 & 250982 \\
\end{tabular}
\end{table}

\subsection{Classification Performance}

Table \ref{table:performance} displays the expected number of weighted signal and background events each model selects per year, as assessed on the final evaluation dataset.
Here the GNN outperforms the physics baseline by identifying 6.3x more signal events while providing a signal-to-noise ratio (SNR) which is 3x improved, also outperforming the CNN.

\begin{figure}[htbp]
\centerline{\includegraphics[scale=0.60]{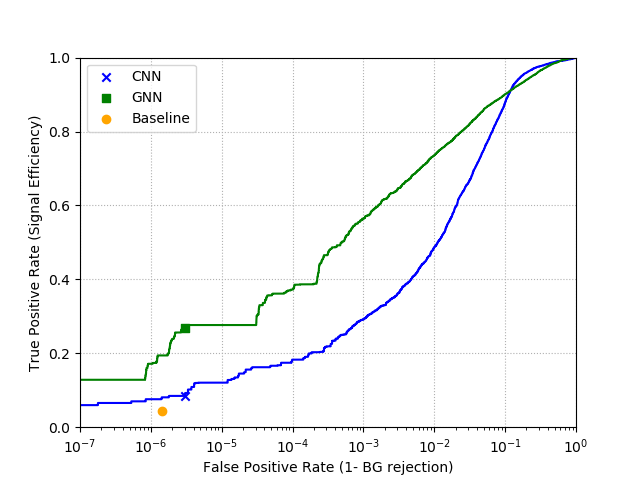}}
\caption{Receiver operating characteristic curve for various methods considered in this paper. The green square and blue X indicate the evaluation point for the GNN and CNN, respectively.}
\label{fig:roczoom}
\end{figure}

Fig. \ref{fig:roczoom} displays receiver operating characteristic (ROC) curves for both the GNN and CNN models.
The physics baseline is directly optimized to maximize the number of signal events identified while attempting to maintain an SNR of 1.0; thus it does not have a ROC curve.
Because there are so many more weighted background events than signal events present, of primary interest is the true positive rate when the FPR becomes small.

\begin{table}[htbp]
 \caption{Performance of several methods}
\label{table:performance}
\begin{tabular}{l*{3}{r}}
& \multicolumn{2}{c}{\# events per year} & \\
\cmidrule(lr){2-3} 
Method & Signal & Background & Signal:Noise \\ 
\midrule
Physics Baseline & 0.922 & 0.934 & 0.987 \\ 
3D CNN & 1.815 & 1.937 & 0.937 \\
GNN & \textbf{5.772} & 1.937 & \textbf{2.980} \\
\end{tabular}
\end{table}


\subsection{Implementation Details}
All code for the GNN model was written in Python using the PyTorch deep learning framework.
Models were trained using Nvidia Tesla P40 GPUs and required approximately 40 hours for 100 epochs of training over 134741 samples, though typically model convergence occurred after just 40 epochs.

\section{Discussion}

The physics baseline result was obtained by optimizing selections on stochasticity in a selected phase space of muon energy proxy and muon zenith angle.  The optimization was performed on the full simulated data sets leading to the final cut level numbers of 0.92 and 0.93 events per year for signal and background, respectively. There are some upgrades to the physics analysis performance level that may increase signal and decrease background. One of them is incorporation of data from the surface IceTop detector and exploitation of its veto capabilities~\cite{TP:neutrino2018}. 

Although the neural network studies were meant to selected events based on the same stochasticity criteria, studies of the selected events found that the networks were also selected starting tracks, where a neutrino interacted in the detector.  Starting tracks are studied by IceCube, but in a different set of diffuse analyses \cite{Aartsen:2018vez,Aartsen:2014muf}.  They were excluded from the current physics baseline analysis, and, in any case, the chosen cuts would strongly discriminate against these events.  A fully parallel treatment of these events would change the comparison between the baseline and machine learning results. 

Despite having small sample sizes, this study shows the possibility of a marked improvement from application of the GNN over the physics baseline and demonstrates a huge potential of this approach for signal classification within the IceCube detector.

\section{Conclusions}
In this work, we studied the application of graph neural networks to the challenging problem of signal detection in the IceCube neutrino observatory. 
We noted the key difficulties within this domain, including the irregular geometry of the detector and the sparse nature of signal it records, and including the large asymmetry between frequency of positive and negative event occurrence.

We presented a graph neural network architecture which allows for adaptive computation by operating on the input signal support. 
Quantitatively, this architecture results in a 3x improvement in signal-to-noise ratio, and is capable of detecting on the order of 6.3x more signal events than traditional physics-based baselines.
While our results are still preliminary, the application of GNNs for the IceCube signal classification task are very promising and we expect future work to further improve the accuracy.

%

\section*{Acknowledgment}
We thank the IceCube Collaboration for their support on this project. This work was supported in part by the National Science Foundation under grant number PHY-1307472 and the U.S. Department of Energy under contract number No. DE-AC02-05CH11231.  This research used computational and storage resources of the National Energy Research Scientific Computing Center (NERSC), a DOE Office of Science User Facility.
FM and MB are supported in part by ERC Consolidator Grant No. 724228 (LEMAN), Google Faculty Research Awards, an Amazon AWS Machine Learning Research grant, an Nvidia equipment grant, a Radcliffe Fellowship at the Institute for Advanced Study, Harvard University, and a Rudolf Diesel Industrial Fellowship at IAS TU Munich. JB and NC are partially supported by the Intel IPCC Program and by the Alfred P. Sloan Foundation.

\bibliographystyle{IEEEtran}
\bibliography{gcn,icecube}

\end{document}